\documentclass[sigconf]{acmart}
\usepackage{graphicx}
\usepackage{subfigure}
\usepackage{makecell}
\usepackage{amsmath}

\usepackage{amssymb}
\usepackage{amsfonts}
\usepackage{mathrsfs}
\usepackage{multirow}
\usepackage{algorithmic}
\usepackage{xcolor}
\newtheorem{definition}{Definition}

\AtBeginDocument{%
  \providecommand\BibTeX{{%
    \normalfont B\kern-0.5em{\scshape i\kern-0.25em b}\kern-0.8em\TeX}}}

\setcopyright{none}             
\renewcommand\footnotetextcopyrightpermission[1]{} 
\begin{document}

\title{SMart: A Multi-source Multi-phase Time Series Representation Transfer Framework}

\author{Fang He, Wang-chien Lee}
\affiliation{\institution{Department of Computer Science and Engineering, \\The Pennsylvania State University, PA, USA}}
\email{Emails: wh27hf@gmail.com, wlee@cse.psu.edu}



\begin{abstract}
Time series representation learning (TSRL) has attracted growing research interests in recent years. Two recent explorations in TSRL are: i) exploiting a transformer-based framework to learn time series; ii) instead of using only the targeted dataset, borrowing time series from other datasets to to facilitate representation transfer. While these two explorations are shown effective, the self-supervised time series recovery task in (i) and the single-source dataset used in (ii) are technically simple and thus can be enhanced with new ideas.
In this work, we propose a new TSRL framework, namely {\em multi-source multi-phase time series representation transfer (SMart)}, which has two novel mechanisms to address the aforementioned deficiencies: 1) a {\em multi-phase recurrence plots recovery task}, in three alternative modes, for guiding the encoder to embed {\em time series dynamics} into the time series representation; and 2) a {\em source dataset selector} to select multiple suitable source datasets to 
supplement the original target dataset for pre-training the TSRL encoder. 
Experimental results show that SMart outperforms several state-of-the-art models 
for time series representation learning, classification and regression
on both uni-variate and multi-variate time series datasets, reducing mean absolute error up to 19.5\% for time series regression, and increasing average accuracy up to 1.34\% for time series classification.
\end{abstract}

\maketitle

\pagestyle{plain}               



\section{Introduction}


A \emph{time series} is a sequence of \emph{observations} over time, where each observation consists of the values of one or multiple variables at a time point. 
Time series data is ubiquitous in the real world, e.g., water levels, human activities and stock prices over time, which may be mined by data mining tasks such as time series classification~\cite{denton2017unsupervised, villegas2017decomposing}, regression~\cite{zerveas2021transformer}, forecasting~\cite{fortuin2018som,gensler2016deep} and clustering~\cite{lei2017similarity, franceschi2019unsupervised}.
To provide general support for various downstream applications, {\em time series representation learning (TSRL)}, aiming to learn representations of input time series for use as general-purpose features, has been widely studied.
Among them, some works focus on learning time series representations for time series of specific types~\cite{saeed2019multi,sarkar2020self,oord2018representation}, while others target on general time series without being limited to certain types~\cite{malhotra2017timenet,lyu2018improving,zerveas2021transformer}, which is also the target of our work.
Recently, Zerveas et al. propose a state-of-the-art TSRL framework, \textit{called Time Series Transformer (TST)}, to learn representations
for general-type time series with a \textit{self-supervised time series recovery task} (\textit{self-recovery} task in short)~\cite{zerveas2021transformer}. 
Specifically, the self-recovery task challenges an encoder in TST to extract features from a masked input time series as its representation, which in turn is fed into a decoder to recover the masked part of the input time series.
While the self-recovery task is shown effective for TSRL, we suggest that adding some useful decoding tasks may further challenge the encoder to enrich the learned representations.
Thus, in this paper, we explore additional self-supervised tasks for TSRL.

Another aspect explored in this work is the issue of data sparsity. Many TSRL frameworks (including TST) face this issue in training the encoder, due to the limited number of time series in a target dataset.
To address the issue, some works exploit the idea of \textit{representation transfer}, i.e., borrowing time series in  additional source datasets to pre-train an encoder for use in TSRL on a target dataset. 
Recently, Fawaz et al. propose DTW-FCN, where a DTW-based similarity metric is designed to find a source dataset most similar with the target dataset to facilitate representation transfer. 
However, the manually designed DTW-based similarity metric may not pick a beneficial source dataset for a target dataset.
Moreover, 
these existing works do not explore \textit{multiple} source datasets which we believe may further benefit representation transfer 
to improve TSRL.

Based on the observations above, we aim to explore the ideas of incorporating new self-supervised tasks for TSRL and using multi-source datasets for pre-training in TSRL. To realize these ideas in a coherent TSRL framework, two research problems arise: 1) the design of a TSRL model with new and complementary self-supervised tasks.

\begin{figure}[t]
\centering
\includegraphics[width=2.6in]{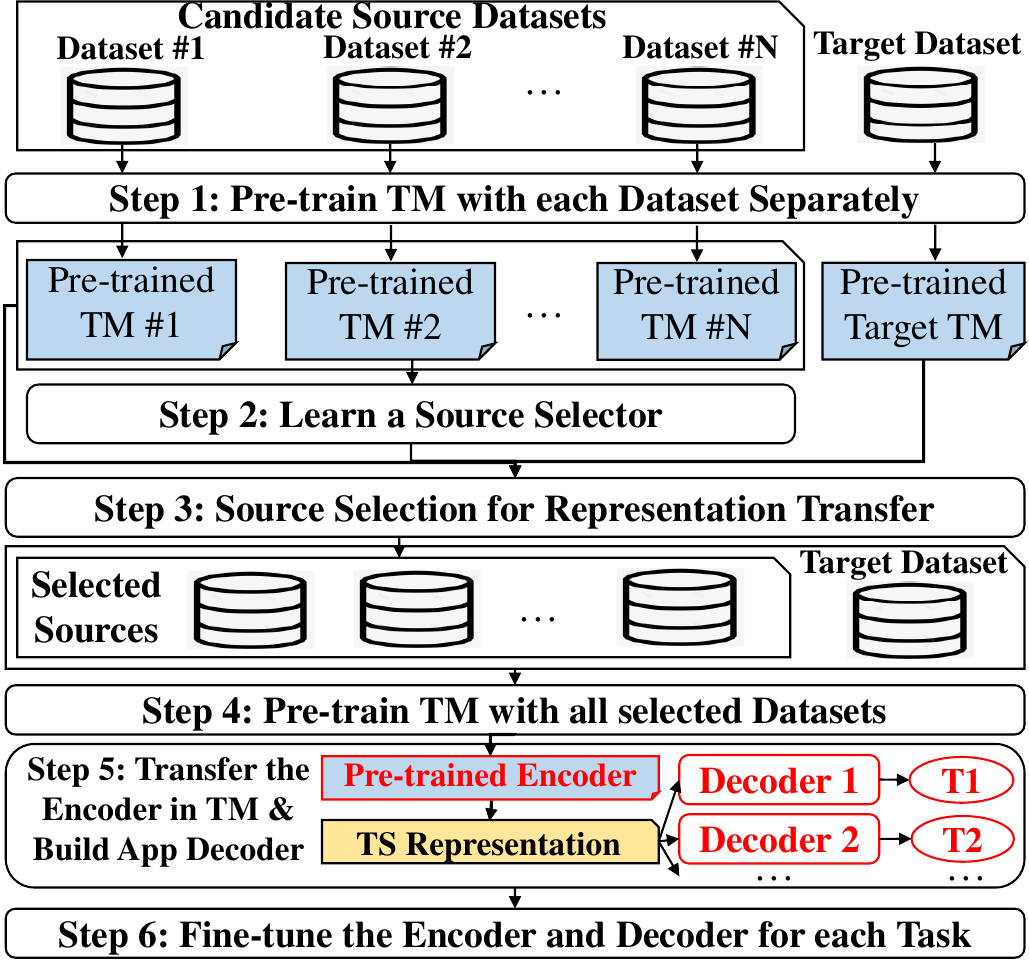}
\caption{SMart Framework}
\label{fig:framework}
\end{figure}

Our approach is to challenge our proposed model to recover the time series dynamics, in addition to the time series itself.
We notice the ability of recurrence plots (RP) to reveal the time series dynamics in terms of how time series status at different time points are correlated with each other, and
thus propose the \textit{Recurrence Plot (RP) Recovery Task} to decode the time series representation to RPs in order to supplement the self-recovery task. 
Owing to the rich multi-granularity time series dynamics embedded in the RPs constructed in different phases, we design three alternative decoding modes to regularize the learned representations in order for them to recover RPs in different phase spaces in different ways; and 2) the selection of (multiple) source datasets which are beneficial to time series representation transfer. Our approach is to formulate the selection of source datasets as a binary classification problem: given a target dataset T and a candidate dataset C to serve as a source of representation transfer, we train a classifier to predict whether C as a source dataset benefits TSRL for T or not, i.e., increase the performance on downstream time series applications (e.g., classification and regression) or not.
Specifically, we propose to learn a neural-network-based Source Selector as the classifier. As such, we can feed each pair of candidate source dataset and the target dataset to the Source Selector to decide whether to select this candidate source dataset for representation transfer.
One issue here is how to represent a (whole) dataset as an input to the Source Selector. Some works use the most representative time series in the dataset as the representation. However, time series in a dataset may vary (e.g., the water level time series recorded in four seasons), making it difficult to find one most representative time series.
To address the issue, we propose a novel idea, i.e., use the \textit{encoder parameters} obtained by pre-training an encoder with the time series dataset as the representation of the whole dataset.
The idea aims to explore the relevance between the ways of extracting meaningful features (represented by the encoder parameters) from target dataset T and the candidate dataset C in order to decide whether to use C as a source dataset in representation transfer for T.
We design a cross-attention mechanism in the Source Selector to capture the relevance between the encoder parameters.

To realize the aforementioned ideas for TSRL, we propose the {\em Multi-\underline{S}ource \underline{M}ulti-Ph\underline{a}se \underline{r}epresentation \underline{t}ransfer (SMart)} framework, which follows the pipeline in Figure~\ref{fig:framework}: 
1) Build a self-supervised Time Series Model (TM) which consists of an encoder and several decoders corresponding to the self-recovery and RP-recovery tasks.
In TM, there are three alternative decoding modes to recover various RPs 
in order to capture multi-granularity time series dynamics;
2) Learn a cross-attention based Source Selector that captures the relevance between encoder parameters of TMs for source selection;\footnote{Encoder parameters of a dataset is obtained by pre-training the TM built in step 1.} 3) Exploit the learned Source Selector to select multiple source datasets for the target dataset; 4) Pre-train a TM with all the selected source datasets and the target dataset; 5) Transfer the encoder in the pre-trained TM in Step 4 to generate representations for each time series in the target dataset and build tailored application decoders to decode these representations to support downstream tasks;\footnote{The application decoder may be a linear layer with a softmax function for time series classification or a linear layer for regression.} and 6) Fine-tune the encoder and train the application decoder together, for each downstream application.

We evaluate SMart against the state-of-the-art i) unsupervised 
TSRL 
frameworks, e.g., TST, DTW-FCN and TS2Vec; and ii) time series classifiers and regressors, such as ROCKET, Rel-CNN and TimeNet.
Experimental results show that SMart outperforms all these baselines in both time series classification and regression.

The major contributions made in this work are as follows.

\begin{itemize}
    \item We propose a novel framework, SMart, for TSRL by integrating the ideas of challenging its encoder to enrich the time series representation in order to recover the dynamics in time series and pre-training encoders with multiple selected source datasets for representation transfer.

    \item We exploit recurrence plots in multiple phase spaces as additional decoding targets to embed the multi-granularity time series dynamics into time series representations.
    
    \item We propose a novel cross-attention based Source Selector to select multiple beneficial source datasets to pre-train the time series encoder for use on the target dataset. We explore the use of encoder parameters as the dataset representation.

    \item Empirical evaluations show that SMart outperforms the state of the arts by significantly reducing the mean absolute error (MAE) by 19.5\% for regression, while competitively improving the accuracy by 1.34\% for classification. 
 
\end{itemize}

\section{Related Work}
In this section, we discuss the existing works for time series classification, regression, and representation learning.

\subsection{Time Series Classification and Regression}
We classify existing works on time series classification and regression in two folds: i) non-deep learning methods, including elastic ensemble (EE)~\cite{lines2015time}, shapelet transform ensemble (STC)~\cite{hills2014classification}, random interval spectral ensemble (RISE)~\cite{flynn2019contract}, time series forest (TSF)~\cite{deng2013time}, proximity forest (PF)~\cite{lucas2019proximity}, HIVE-COTE~\cite{lines2016hive}, TS-CHIEF~\cite{shifaz2020ts} and ROCKET~\cite{dempster2020rocket}, which design features (e.g., elastic similarity measures, shapelet-based features and interval-based features) and ensemble a set of models for time series classification and regression; and ii) deep learning models, including InceptionTime~\cite{ismail2020inceptiontime}, ResNet~\cite{fawaz2018data} and Rel-CNN~\cite{9809831}, which mostly stack convolutional layers to automatically extract features for time series mining tasks. These works, however, are complex and hard to scale to datasets with many time series samples and long time series. Moreover, it is not easy to adapt these works for other time series applications.

\subsection{Time Series Representation Learning}
Time Series Representation learning (TSRL) has attracted growing research interests in recent years~\cite{woo2022cost, zerveas2021transformer,yue2022ts2vec,eldele2021time,li2022towards,saeed2019multi, lei2017similarity}. Many works focus on learning representations for time series for a specific task, e.g., human activity recognition~\cite{saeed2019multi}, emotion recognition~\cite{sarkar2020self} and speech time series~\cite{oord2018representation}. However, these works are not designed for general-type time series. 
In addition, Lei et al. propose to learn similarity-preserving representations for time series by capturing the similarity between two time series in terms of Dynamic Time Warping (DTW) in their representations~\cite{lei2017similarity}. However, these similarity-preserving representations do not show advantages for time series applications such as classification and regression.

Recently, several general-type TSRL frameworks are proposed based on Recurrent Neural Networks, dilated CNN or Transformers. Among them, Malhotra et al. propose a multi-layered RNN based time series autoencoder, to reconstruct the time series in a sequence-to-sequence manner~\cite{malhotra2017timenet}.
 Lyu et al. propose a multi-layered LSTM with an attention mechanism for unsupervised time series representation learning~\cite{lyu2018improving}.
Yue et al. propose a dilated CNN based TS2Vec framework via contrastive learning for TSRL~\cite{yue2022ts2vec}.
Owing to the strength of Transformer in extracting long-term features, Zerveas et al. propose TST, 
a transformer-based model that encodes a masked input time series to its representation and then decodes the representation to recover the masked part of the input time series in order to capture useful features in the representation~\cite{zerveas2021transformer}. However, these SOTA TSRL frameworks train the encoder using the target dataset only, without exploiting additional datasets to learn more knowledge for representation transfer.
In this trend, Fawaz et al. propose a DTW-based similarity measurement between two datasets, aiming to select a source dataset similar to the target dataset to pre-train a fully convolutional network, which in turn is fine-tuned for applications on the target dataset~\cite{fawaz2018transfer}. However, it does not exploit the potential of multiple source datasets for TSRL.

\section{Preliminaries}
In this section, we first formulate the research problem.
Next, we provide the background on {\em recurrence plots} and the idea of source selection, which lead to two main improvements in this research. Finally, we analyze issues in realizing our approaches.

\subsection{Problem Formulation}

\begin{definition} {\bf Time series}.  
A time series $TS$ consists of a series of values sampled in a fixed time gap, i.e.,
$ TS = [\vec{v_1},\vec{v_2},...,\vec{v_l}]$ where $l$ is the time series length, $\vec{v_i}=(v_i^1, ..., v_i^n)^T$ is a vector of values recording the variables observed at the time point $i$, $n$ is the number of variables, and $v_i^j$ is the value of the $j$-th variable at time point $i$.
\end{definition}

Next, we define the task of \textit{time series representation learning (TSRL)} which is the ultimate problem we aim to tackle in this work. 

\begin{definition}
{\bf Time Series Representation Learning (TSRL)}. 
Given a time series dataset $\mathbf{D}$ = $\{{TS}_{i}\}^{|D|}_{i=1}$, the time series representation learning task is to learn an encoder that maps a time series, e.g., $TS_i$, to its latent representation $z_i\in \mathbb{R}^d$
in support of downstream tasks in time series applications (e.g., classification and regression). 

\end{definition}

In this paper, we design a novel framework, called {\em SMart} for TSRL, with the ideas behind the tasks of recurrent plots recovery and source selection. We introduce recurrence plots in Section 3.2, which serves as the foundation for the proposed recurrent plot
recovery task. We discuss source selection in Section 3.3.

\subsection{Recurrence Plots} \label{RP_background}
Recurrence plots, in classical phase space reconstruction theory, describe time series dynamics~\cite{eckmann1995recurrence,marwan2007recurrence,kennel1992determining}. Basically, 
a phase point in a multi-dimensional phase space denotes the status of time series at a time point, and a recurrence plot records how the phase points (i.e., the time series status) at different time points correlate with each other (i.e., the time series dynamics). 

\begin{figure}[t]
\centering
\includegraphics[width=3in]{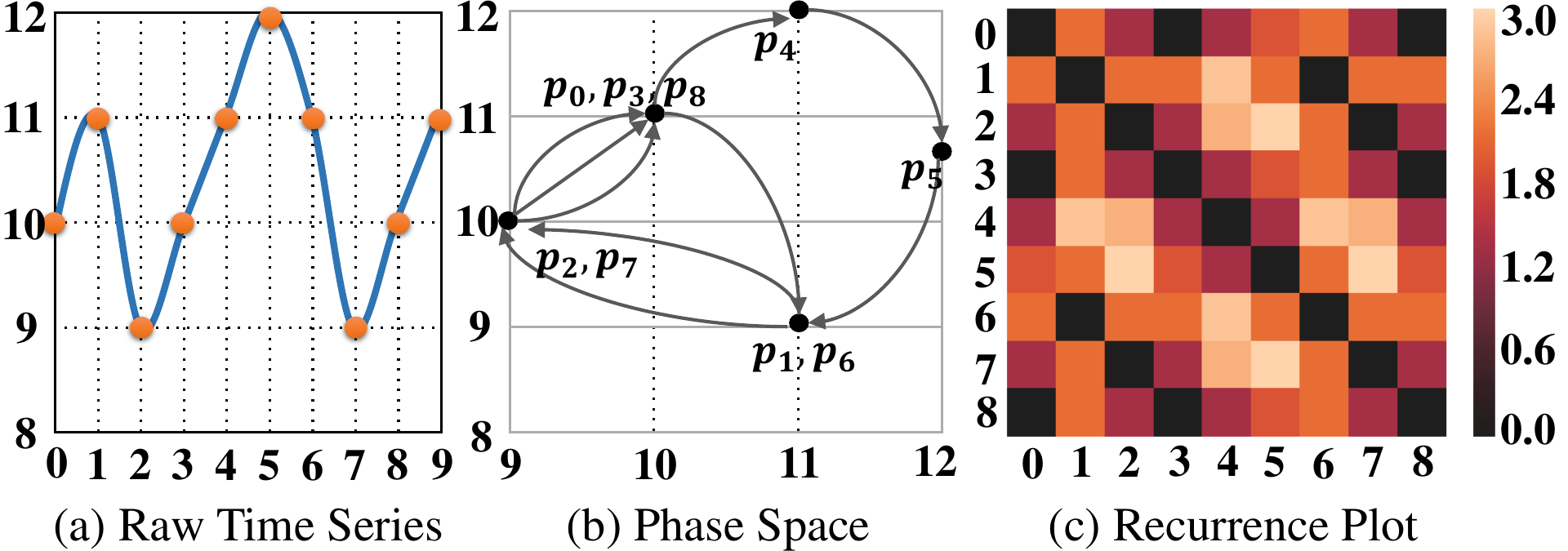}
\caption{Recurrence Plot Construction}
\label{fig:recurrenceplot}
\end{figure}

Phase space reconstruction is controlled by two hyper-parameters, the phase space dimension factor $d$ and a time delay $\tau$.
Given $d$, $\tau$ and a time series $TS = [\vec{v_1},\vec{v_2},...,\vec{v_l}]$, for each time point $t$ ($t=1,2,...,l-(d-1)\tau$), we generate a phase point, $p_t$ = ($\vec{v_t}^T$, $\vec{v}^T_{t+\tau}$, ..., $\vec{v}^T_{t+(d-1)\tau}$)$^T$$\in R^{d\cdot n}$, where $n$ is the number of variables in $TS$. 
By linking the phase points of $t=1,2,...,(l-(d-1))\tau$ one by one, we gain a trajectory in the phase space starting from $p_1$, going through $p_2$, $p_3$, etc., and ending with $p_{l-(d-1)\tau}$. 
Based on the phase space reconstruction theory, the phase space preserves the topology of the system of variables in the time series, and the reconstructed trajectory shows similar dynamics as the raw time series~\cite{mcguire1997recurrence}. 
Thus, a recurrence plot $RP_{d,\tau}$ (a two-dimensional matrix) is derived from the trajectory to describe, for each phase point $p_t$, how other phase points are far away from $p_t$. Specifically, the value in the $i$-th row and the $j$-th column of $RP_{d,\tau}$ is derived by $RP_{d,\tau}(i,j)=f(p_i,p_j)$, where $f$ is the Euclidean distance in the phase space.

As an example, Figure \ref{fig:recurrenceplot}(a) shows a uni-variate time series $TS=[v_0=10, v_1=11, ..., v_9=11]$. Figure \ref{fig:recurrenceplot}(b) shows the two-dimension phase space reconstructed from $TS$ with $d=2$ and $\tau=1$, where the phase point $p_i$ is ($v_i$, $v_{i+1}$) and the trajectory (linked by the arrows) goes through all the phase points in the temporal order. Figure \ref{fig:recurrenceplot}(c) shows the recurrence plot $RP_{2,1}$ generated from Figure \ref{fig:recurrenceplot}(b), where $RP_{2,1}(i,j)=Euclidean(p_i,p_j)$ capturing how the phase point $p_i$ correlates to $p_j$. For example, in \ref{fig:recurrenceplot}(c), the value in the 0-row and 8-column is close to 0, i.e., the time series status at time point 8 is similar to that at time point 0, suggesting a potential periodicity.

\subsection{Source Selection}
Inspired by advances on transfer learning in natural language processing~\cite{devlin2018bert,brown2020language,yang2019xlnet} and computer vision~\cite{he2016deep,simonyan2014very,krizhevsky2017imagenet} that exploits huge amount of data for pre-training an encoder which can be transferred to serve as an initial encoder and fine-tuned along with tailored decoders in relevant applications, SMart follows this transfer learning paradigm to address the issue of training data sparsity.
Accordingly, we define \textit{time series representation transfer} as follows. 

\begin{definition}
{\bf Time Series Representation Transfer (TSRT)}. 
Given a set of time series datasets, i.e., $Set(D)=\{D_1, D_2, ..., D_N\}$ and a target dataset $D_{target}$, time series representation transfer explores some source datasets in $Set(D)$ to pre-train an encoder that is transferred as the initial encoder to assist TSRL in $D_{target}$.
\end{definition}

It is challenging to decide which datasets in $Set(D)$ may help pre-train the encoder. We define the source selection as follows.

\begin{definition}
{\bf Source Selection}. 
Datasets in $Set(D)$ are called \textit{candidate source datasets}. In TSRT, the datasets selected from $Set(D)$ for encoder pre-training are called \textit{source datasets}. \textit{Source selection} is the process of selecting source datasets from candidate source datasets.
\end{definition}

\subsection{Problem Analysis}
As mentioned earlier, the SMart incorporates two tasks: recurrence plots recovery (i.e., \textit{RP-recovery}) and source selection for TSRL. The goal of RP-recovery is to exploit additional decoders to recover multiple recurrence plots in order to challenge/guide the encoder to capture time series dynamics of different granularity. However, given recurrent plots of different granularity, what kinds of decoder designs in SMart is able to effectively achieve the goal of RP-recovery? Recent TSRL frameworks based on transformer architecture usually have multiple layers of encoders, which thus generate multiple sets of time series representations, correspondingly. Should we focus on training the representations generated in the last encoding layer for RP-recovery or taking into account the representations in different layers? To address this issue, we propose three alternative designs, namely {\em Recover-At-The End (Rate)}, {\em Recover-in-Sequence(RS)} and {\em Recover-in-Progressive-Way (RPW)}, which along with our approach for decoding, are detailed in Section~\ref{sect:rpr}.

On the other hand, source selection is a key issue in TSRT. Not all the candidate source datasets are useful for pre-training the encoder. Existing works actually show that transfer learning from some specific time series dataset may be harmful~\cite{fawaz2018transfer}.
To tackle the issue of source selection, our idea is to treat it as a binary classification problem, and learn a neural-network based source selector as the classifier to predict the feasibility (i.e., positive or negative) of pre-training the encoder for TSRL with a certain source dataset, given a target dataset. 
As such, we can feed a candidate source dataset and the target dataset to the source selector for source selection. We detail our ideas and implementations Section~\ref{sect:MSTL}.

\section{Framework Design}
In this section, we present the design of the SMart framework.
We first briefly recall the time series transformer, which is used as the backbone model in SMart.
Then we extend it by adding a recurrence plots recovery module. Finally, we discuss the multi-source transfer learning in SMart, with the design of Source Selector detailed.

\subsection{SMart Framework}
\begin{figure}[t]
\centering
\includegraphics[width=2.5in]{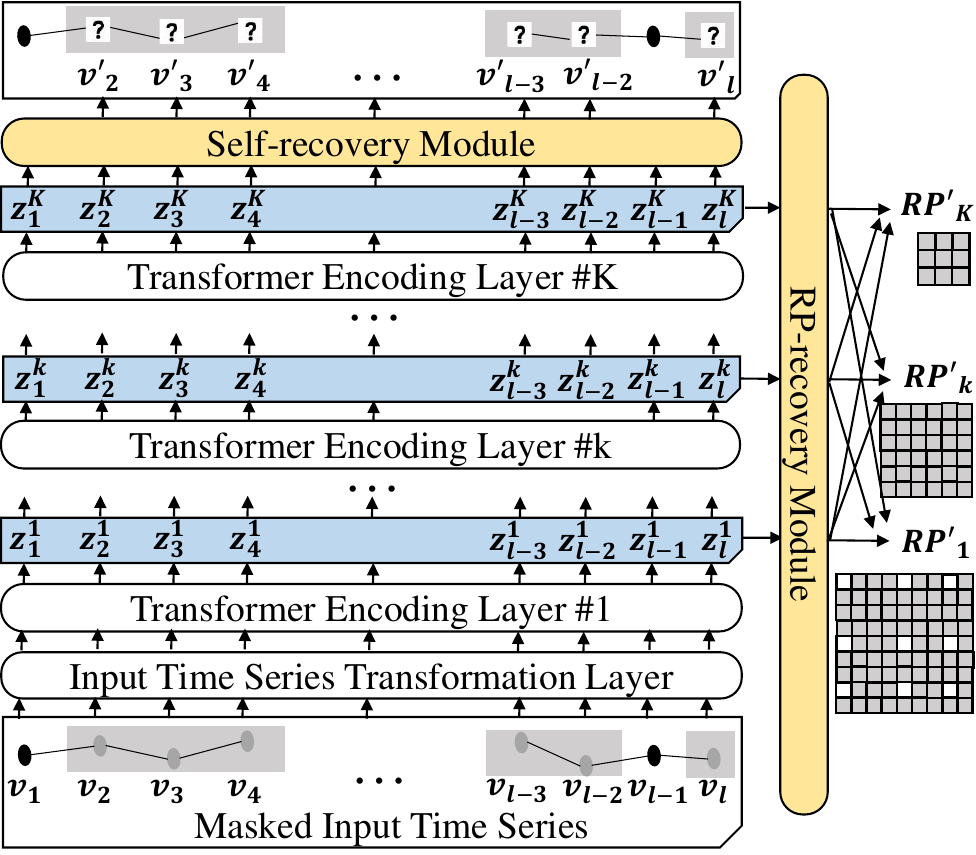}
\caption{Time Series Model with both Self and RP Recovery}
\label{fig:recurrenceplotrecovery}
\end{figure}

In this paper, we propose new framework, namely {\em Multi-\underline{S}ource \underline{M}ulti-Ph\underline{a}se \underline{r}epresen-tation \underline{t}ransfer (SMart)} for time series representation learning.
As shown in Figure~\ref{fig:framework}, given a target dataset and a set of candidate source datasets, SMart learns representations of the time series in the target datasets in the following steps: 1) Build a Time Series Model (TM) which follows an encoding-decoding paradigm, and train it with each dataset (including candidate and target datasets), respectively. To embed time series dynamics into time series representations, TM (shown in Figure~\ref{fig:recurrenceplotrecovery}) combines the base time series model (to be introduced in Section 4.2, with only the self-recovery module) and the recurrence plot recovery module (to be introduced in Section 4.3). Both the self-recovery module and the RP-recovery module (in yellow) are decoders in TM, aiming to challenge the encoder (layers in white, including the input time series transformation layer and the transformer encoding layers) to encode essential information and synamics in time series. 2) Learns a Source Selector (to be introduced in Section 4.4) for source selection, using the (target, candidate) pairs of TMs trained in step 1 as the training data.
3) Use the learned Source Selector to decide source datasets suitable for the target dataset from the candidate datasets. More specifically, we pair each candidate source dataset and the target dataset as input to the Source Selector to decide whether the candidate source dataset is good to use for the target dataset; 
4) Pre-train a TM with all the selected source datasets and the target dataset; 5) Transfer the encoder parameters in the TM trained in step 4 to serve as the initial encoder for various applications. The output of the encoder is fed to task decoders (e.g., a classifier or a regressor) for downstream applications. 
and 6) Fine-tune the encoder while training the application decoders with the training set of the target dataset, and evaluate the applications on the test set.

\subsection{Backbone: Time Series Transformer}
SMart adopts time series transformer~\cite{zerveas2021transformer} as its backbone time series model (TM).\footnote{We exploit transformer as the backbone model since it is reported to outperform other encoder structures such as ResNet, LSTM and dilated CNN~\cite{zerveas2021transformer}.} First, time series data is normalized and masked. A time series $TS = [\vec{v_1},\vec{v_2},...,\vec{v_l}]$, where $\vec{v_t}=(v_t^1, ..., v_t^n)^T$, 
is masked via $TS \odot \textbf{M}$ where $\textbf{M}$ is a binary mask matrix $\in \{0,1\}^{n\times l}$ created independently for each $TS$ and epoch. 
We denote the masked input time series as $V=TS\odot \textbf{M} = [v_1, v_2, ..., v_l] \in \mathbf{R}^{n\times l}$, where $v_t\in \mathbf{R}^n$ is the input vector at time point $t$.
As shown in Figure~\ref{fig:recurrenceplotrecovery} (but without the RP-recovery module), each value vector $v_t$ is fed to an \textit{input time series transformation layer} (with a projection and positional encodings added) to form the initial representation of that time point, i.e., $x_t \in \mathbf{R}^{d_0}$, where $d_0$ is the transformer model dimension.
$\{x_t\}$ are then fed to $K$ stacked transformer encoding layers to generate the time series representation. 
We denote the output representation at time point $t$ by the encoding layer $\#i$ as $z^i_t$ (as shown in the blue boxes). The representations output by the last ($\#K$) encoding layer, i.e., $\{z^K_1, z^K_2, ..., z^K_l\}$, are concatenated as the representation of the whole time series, denoted as $z\in\mathbf{R}^{d_0\cdot l }=[z^K_1 || z^K_2 || ...|| z^K_l]$. 
To train the encoder, a self-recovery task is exploited to enforce the learned representation of each time point to capture information needed for recovering the raw value vector at the same time point (only the losses of the recovered masked parts are considered). 
More specifically, we recover $v_t$ by $v^{\prime}_t=W_oz^K_t+b_o$, where $W_o$ and $b_o$ are learnable parameters, and measure the loss by mean square error of the masked part between $\{{v}^\prime_t\}$ and $\{\vec{v}_t\}$ as follows.
\begin{equation}
\setlength{\abovedisplayskip}{3pt}
\setlength{\belowdisplayskip}{3pt}
    \mathcal{L}_{self} = \frac{1}{N_\textbf{M}}\sum^l_{t=1}\sum^n_{q=1}({v^\prime_t}^q-v^q_t)^2\cdot (1-\textbf{M}_{q,t})
\end{equation}
where $\textbf{M}_{q,t}$ is the mask value for the $q$-th variable at the time point $t$, and $N_\textbf{M}$ is the number of $0$'s in the mask matrix $\textbf{M}$ (i.e., the number of values that are masked). The self-recovery task is to train the encoder by minimizing the sum of $\mathcal{L}_{self}$ for all time series.

\subsection{Recurrence Plot (RP) Recovery}\label{sect:rpr}

In this section, we propose the recurrence plot recovery (RP-recovery) task to further enrich the time series representation by adding decoders to recover the RPs additionally.
As introduced earlier,
a RP is generated from a phase space with the pre-set $d$ and $\tau$; RPs with different $d$ reveal the time series dynamics in different granularity. 
Let a RP with parameter $d$ be denoted as $RP_{d}$ (i.e., $RP_{d,\tau=1}$).\footnote{For simplicity, we only explore the fundamental recurrence plots with $\tau=1$ in this paper. The exploration of other recurrence plots are left to the future work.} 
The goal of the RP-recovery task is to decode the representation to recover $RP_1$, ..., $RP_{K}$ to capture dynamics described in those plots.

Here we first analyze some interesting characteristics of $\{RP_{d} | d=1,2,...,K\}$. i) $RP_{d}$ of a larger $d$ is more challenging to recover than one with a smaller $d$, 
because the encoder may lack information of intermediate RPs to generate the representation of a phase point w.r.t. a larger $d$ so as to recover $RP_d$. ii) RPs of smaller $d$s provide hints to recover the RPs of larger $d$s. For example, when we observe that in $RP_1$, the phase point at time point $t_1$ (i.e., $p^{d=1}_{t_1}$) is shown to be close to $p^{d=1}_{t_2}$ at $t_2$ and meanwhile $p^{d=1}_{t_1+1}$ is close to $p^{d=1}_{t_2+1}$, we can infer that in $RP_2$, $p^{d=2}_{t_1}=(p^{d=1}_{t_1}|| p^{d=1}_{t_1+1})$ should be close to $p^{d=2}_{t_2}=(p^{d=1}_{t_2}||p^{d=1}_{t_2+1})$. iii) the representations output by different encoding layers, i.e., $\{{z^k_1, ..., z^k_l}|k=1,2,...,K\}$ have similar characteristics with $\{RP_d\}$. 
Inspired by these observations, the RP-recovery module is designed to decode the representations generated by different encoding layers to recover the RPs in multiple phase spaces (shown in Figure ~\ref{fig:recurrenceplotrecovery}).

\begin{figure}[t]
\centering
\includegraphics[width=2in]{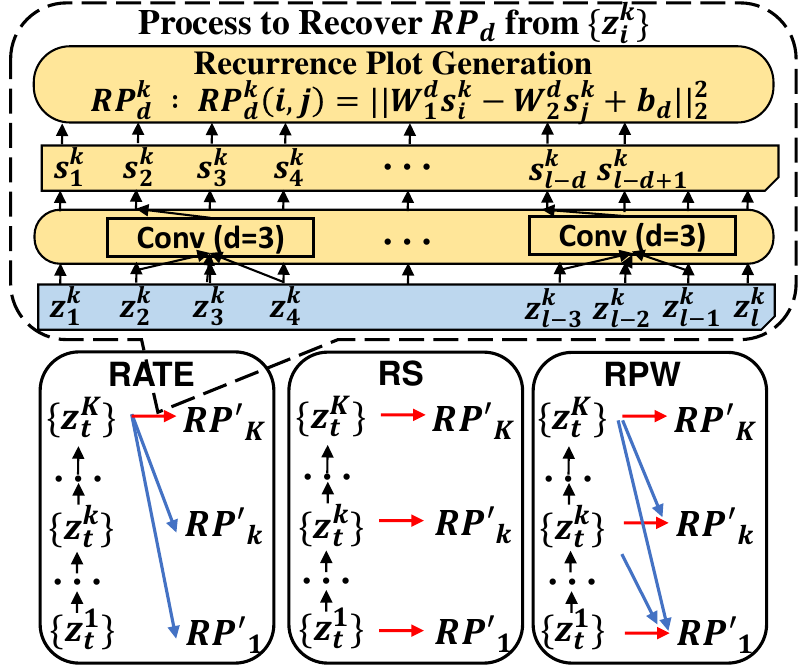}
\caption{RP-recovery Module in alternative modes}
\label{fig:modes}
\end{figure}

We first discuss how we decode the representations output by the encoding layer $\#k$ (i.e., {$z^k_1, z^k_2, ..., z^k_l$}) to $RP_d$. 
The idea is to first general phase point representations from representations of different time points, and then generate the RP from phase point representations.
As shown in Figure~\ref{fig:modes}, $\{z^k_1, z^k_2, ..., z^k_l\}$ are first fed to a convolution layer to aggregate the representations of time points as the representations of phase points. Specifically, we generate the phase point representation at $t$ by $s^k_t = conv_d({z^k_{t}, z^k_{t+1}, ..., z^k_{t+d-1}})$, where $conv_d(\cdot)$ represents a 1D-convolution operation with kernel size as $d$.
Next, we feed the pair of phase point representations at time point $i$ and $j$ to a recurrence plot generation layer to predict the distance between the two phase points in $RP_d$ as follows.
\begin{equation}
\setlength{\abovedisplayskip}{3pt}
\setlength{\belowdisplayskip}{3pt}
    RP^k_d(i,j) = || W^d_1s^k_i-W^d_2s^k_j+b_d ||^2_2, 1\leq i,j\leq l-d+1
\end{equation}
where $W^d_1$, $W^d_2$ and $b_d$ are the learnable matrices and bias, and $RP^k_d(i,j)$ is the predicted value in the $i$-th row and $j$-th column in $RP_d$. To facilitate training, we use the mean square error between $RP^k_d$ and $RP_d$ as the loss function, but we only count the error for the masked values (in grey color) in $RP_d$ (i.e., to calculate the value in $RP_d$, we must know some masked values in the raw time series). Formally, the loss w.r.t. the encoding layer $k$ and $RP_d$ is as follows.
\begin{equation}\small
\setlength{\abovedisplayskip}{3pt}
\setlength{\belowdisplayskip}{3pt}
    \mathcal{L}^{k,d}_{rp} = \sum_i\sum_j \frac{(RP^k_d(i,j)-RP_d(i,j))^2}{N^d_{masked}}
    (1-\prod^n_{q=1} (\prod^{i+d-1}_{t=i}\textbf{M}_{q,t}\prod^{j+d-1}_{t=j}\textbf{M}_{q,t}))
\end{equation}
where $\textbf{M}$ is the mask and $N^d_{masked}$ is No. of masked values in $RP_d$.

Once $\mathcal{L}^{k,d}_{rp}$ is defined, the next step is to design how to select the $(k,d)$ pairs, i.e., decode which layer's representations to which recurrence plots, for the RP-recovery module. 
We classify all RP-recovery tasks (each indicated by a $(k,d)$ pair) into two categories: i) {\em challenging recovery,} i.e., $(k, k)$ for $k=1,2,...,K$, to decode representations output by the encoding layer $\#k$ to the currently most challenging RP (shown as red arrows in Figure~\ref{fig:modes}); and ii) {\em comprehensive recovery}, i.e., $(k,d)$ for $1\leq d<k\leq K$, to decode representations output by encoding layer $\#k$ to comprehensively recover all lower RPs (shown as blue arrows in Figure~\ref{fig:modes}) in order to capture multi-granularity time series dynamics.
Note that challenging recovery from bottom encoding layers (with smaller $k$s) may regularize the representations and provide guidance (due to the aforementioned characteristics) to assist the challenging recovery from upper encoding layers to capture RPs of larger $d$ (which provide more precise time series dynamics~\cite{mcguire1997recurrence}).
In SMart, we propose three alternative modes as described below (also see Figure~\ref{fig:modes}).

\noindent\textbf{Recover-At-The-End (RATE).} An intuitive idea is to decode the time series representations output by the encoding layer $\#K$ to recover all the $RP$s in order to capture time series dynamics of different granularity. 
The total loss is calculated as $\mathcal{L}_{RATE} = \sum^K_{d=1}\mathcal{L}^{K,d}_{rp}$.

\noindent\textbf{Recover-in-Sequence (RS).} Inspired by the observed characteristics of $\{RP_d\}$, we recover the $RP$s of different $d$s sequentially from the easiest to the most difficult one.
Specifically, we decode representations output by each encoding layer with a challenging recovery.
As such, \textbf{RS} divides the encoding process of $RP_K$ into several steps to reduce the difficulty in each step, which makes the encoder training easier. We have the total loss $\mathcal{L}_{RS} = \sum^K_{d=1}\mathcal{L}^{d,d}_{rp}$.

\noindent\textbf{Recover-in-Progressive-Way (RPW).} To explore ideas in both \textbf{RATE} and \textbf{RS}, we decode representations output by each encoding layers with both challenging and comprehensive recovery.
As such, the final representations encode information in the RPs of all $d$s with comprehensive recovery, while the earlier challenging recovery tasks on lower encoding layers provide guidance for later challenging recovery tasks and regularize the encoding process. Formally, we have the total RP-recovery loss $\mathcal{L}_{RPW} = \sum^K_{k=1}\sum^k_{d=1}\mathcal{L}^{k,d}_{rp}$.

By adding the RP-recovery loss to $\mathcal{L}_{self}$, SMart exploits both the self-recovery and RP-recovery task to train a TM.
Experimentally, we evaluate SMart in all these three modes and compare the impacts of challenging and comprehensive recovery in Section 5.

\subsection{Source Selection} \label{sect:MSTL}

A remaining issue in SMart is the design of the Source Selector, which selects multiple source datasets to facilitate time series representation transfer. 
The function of the Source Selector can be reduced to 
predicting the feasibility (positive or negative) of using a given candidate source dataset 
to supplement the target dataset. 
Intuitively, we model the Source Selector as a binary classifier.

Instead of directly feeding the candidate and targeted datasets to  
Source Selector, we explore an interesting idea --- use parameters of the first transformer encoding layer in the TM pre-trained with the dataset. We argue that these parameters capture information regarding how to extract meaningful features from the dataset, and thus may serve as a good representation of the whole dataset. Therefore, datasets are pre-processed to obtain encoder parameters as their dataset representation for use in source selection.~\footnote{We empirically examine and validate this idea in Section 5.2.3.}

To train a Source Selector, we assume a pool of datasets available for training and prepare training entries in the form of (A, B, label), where A and B are datasets from the training pool.\footnote{Note that A and B serve the roles of candidate and target, respectively. We use these symbols under the context within Source selector to avoid confusion with C and T used under general context of SMart.} 
Let A and B be properly split into a training set and a testing set, respectively. We prepare the label by:
a) train the time series model (TM) using only the training set of B to get $encoder_B$; and b) train the TM using both training sets of A and B to get $encoder_{AB}$. 
We then evaluate $encoder_B$ and $encoder_{AB}$ (by further fine tuning and appropriate decoding; to be introduced later) on the testing set of dataset B for time series classification and regression. By comparing the performance of $encoder_B$ and $encoder_{AB}$, we set the label as 1 
if $encoder_{AB}$ outperforms $encoder_B$ on both tasks of classification and regression, and 0 otherwise. Accordingly, we enumerate pairs of datasets in the training pool for different roles to prepare the training data for Source Selector.


\begin{figure}[t]
\centering
\includegraphics[width=2.2in]{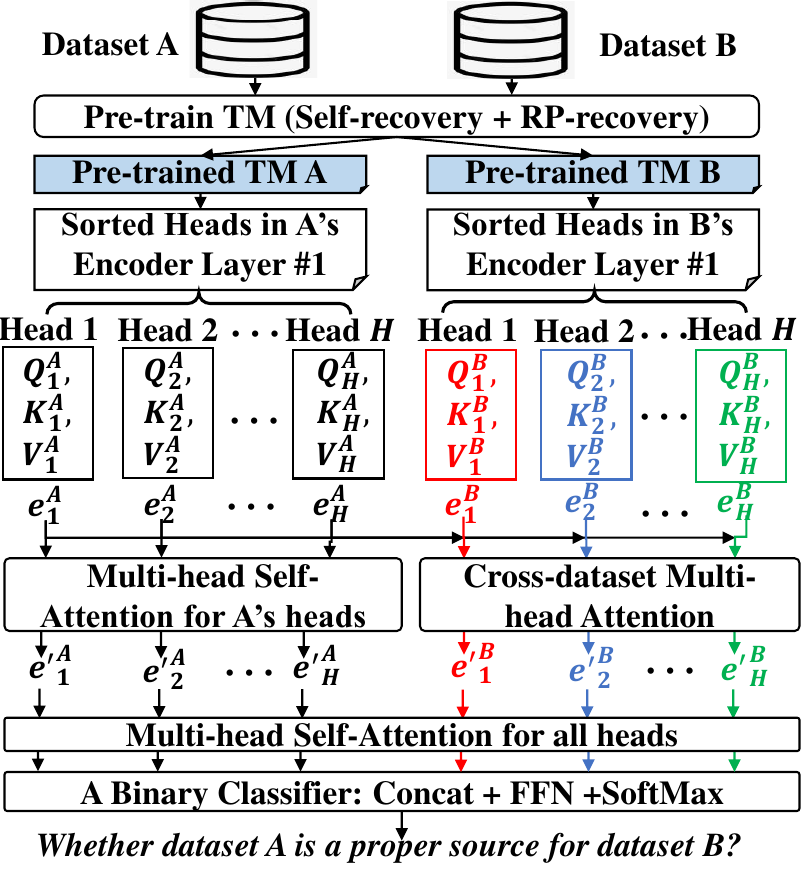}
\vspace{-0.1in}
\caption{Source Selector}
\label{fig:sourceselector}
\vspace{-0.2in}
\end{figure}

Design of the Source Selector is illustrated in Figure~\ref{fig:sourceselector}. 
First, we pre-train the TM with dataset A (to get $TM_A$) and dataset B (to get $TM_B$), respectively.
As discussed, we use the multiple heads in the self-attention layer of the encoding layer $\#1$ as the dataset representation. 
We denote the number of heads by $H$. For the $h$-th head in $TM_A$, the parameters include three projection matrices, i.e., $Q^A_h$, $K^A_h$ and $V^A_h$ for generating the query, key and value vectors, respectively.
Thus, we represent dataset A by $\{Q^A_h\}$, $\{K^A_h\}$, and $\{V^A_h\}$ ( $h=1,2,...,H$) and represent dataset B similarly.
However, when we train the TM with a dataset, we do not specify the order of heads, i.e., we may get another $TM^\prime_A$ when training TM with A, whose heads are in a permuted order of $TM_A$.
To address this issue, for each head, we concatenate its flattened three projection matrices to produce its initial embedding, and sort the heads by the l2-norm of its embedding in increasing order. 
As such, we fix an order for different heads to train Source Selector consistently.

We denote the sorted initial head embedding of the $h$-th head for dataset A as $e^A_h$ and denote that for dataset B as $e^B_h$.
We first project these embeddings into a latent space, with three projection matrics (denoted as $Q$, $K$ and $V$) as follows.
\begin{equation}
\setlength{\abovedisplayskip}{3pt}
\setlength{\belowdisplayskip}{3pt}
    q^{D}_h = Qe^D_h, k^{D}_h = Ke^D_h, v^{D}_h = Ve^D_h, \  h=1,2,...,H \& D\in\{A,B\} 
\end{equation}
To generate the new head embedding for a head in $TM_A$, we follow the basic self-attention mechanism to capture the impact of other heads on that head with attention scores to aggregate head embeddings with the normalized attention scores as weights. Formally, for Head $h$ in $TM_A$, we generate its new embedding ${e^\prime}^A_h$ as follows.
\begin{equation}
\setlength{\abovedisplayskip}{3pt}
\setlength{\belowdisplayskip}{3pt}
    \alpha^A_{j,h} =q^A_hW^A_\alpha k^A_j,     \beta^A_{j,h} = \frac{exp(\alpha^A_{j,h})}{\sum^H_{j=1}exp(\alpha^A_{j,h})}, {e'}^A_h = \sum^H_{j=1}\beta^A_{j,h}v^A_j
\end{equation}
where $W_\alpha$ is the learnable transformation matrix to calculate the attention score $\alpha^A_{j,h}$, and $\beta^A_{j,h}$ is the normalized attention score from head $j$ to head $h$.
To generate the new embedding of Head h for dataset B, we exploit a cross-attention mechanism to let the heads for dataset B to "observe" the heads for dataset A to capture their correlations.
More specifically, we capture the impact of Head $j$ in $TM_A$ to the Head $h$ in $TM_B$ to be $\alpha^B_{j,h}$, normalize the impacts, and generate the new embedding of Head $h$ in $TM_B$, i.e., ${e^\prime}^B_h$, by aggregating the head embeddings with normalized impacts as follows.
\begin{equation}
\setlength{\abovedisplayskip}{3pt}
\setlength{\belowdisplayskip}{3pt}
    \alpha^B_{j,h} =q^B_hW^B_{\alpha} k^A_j,
    \beta^B_{j,h} = \frac{exp(\alpha^B_{j,h})}{\sum^H_{j=1}exp(\alpha^B_{j,h})}, {e'}^B_h = \sum^H_{j=1}\beta^B_{j,h}v^B_j
\end{equation}
Next, we feed both $\{{e^\prime}^A_h\}$ and $\{{e^\prime}^B_h\}$ to another multi-head self-attention layer to derive the new embedding for each head, denoted as $\{{e^*}^A_h\}$ and $\{{e^*}^B_h\}$, respectively. Finally, we concatenate these embeddings to feed to a two-layer feed forward network with a softmax layer for binary classification to predict whether to select dataset A as a source dataset for dataset B or not.

\section{Performance Evaluation}
In this section, we first evaluate the proposed cross-attention based Source Selector and then evaluate the SMart framework for both time series classification and regression. 

\noindent
{\bf Research questions.} We attempt to answer the following research
questions through the experiments and analyses:
\begin{itemize}
    \item \textbf{RQ1}. What about the effectiveness of the Source Selector design? Do the first-layer encoder parameters serve well as the representation of a time series dataset, comparing against conventional representatives of datasets? (Sect. 5.2)
    \item \textbf{RQ2}. What about the effectiveness and efficiency of SMart against the baseline methods? (Sect. 5.3.2)
    \item \textbf{RQ3}. How do various components in SMart, such as the different pre-training tasks and source selection strategies affect the performance of SMart for TSRL? (Sect. 5.3.3)
    \item \textbf{RQ4}. How about different RP recovery modes? (Sect. 5.3.4)
    \item \textbf{RQ5}. How about adapting SMart trained with uni-variate time series datasets for multi-variate datasets? (Sect. 5.4)
\end{itemize}

\subsection{Datasets}
Here we introduce the time series datasets used in the evaluation.

\noindent \textbf{Bake-off uni-variate datasets}. We collect the 85 bake-off uni-variate datasets \cite{bagnall2017great} that are commonly used for time series classification evaluation. For time series regression, we modify those datasets by removing the value at the last time point from each time series. The removed value serves as the ground truth for regression.

\noindent \textbf{UEA multi-variate datasets}. We collect 9 labelled multi-variate datasets from the UEA archive~\cite{bagnall2018uea} with for evaluation of time series classification. For evaluation of time series regression, we remove the value of the last variable (consistent in each dataset) at the last time point from each raw time series and use it as the ground truth.

\subsection{Evaluation of Source Selector}
We evaluate the Source Selector on Bake-off uni-variate datasets. 

\subsubsection{Experiment Setup}\label{source-selector-evaluation}
We randomly split the 85 bake-off datasets into 17 groups, each of which has 5 datasets. Then we randomly pick 2 groups as the testing groups, 2 groups as the validation groups, and the rest 13 groups as the training groups (forming the pool of training datasets for the Source Selector).
To train the Source Selector, we enumerate all combined pairs of datasets from the training groups and obtain the corresponding label as discussed in Section 4.4. We adopt the Adam optimizer for training, with the learning rate = 0.001. 
For validation, we pair each dataset from the training groups with each dataset from the validation groups to generate 65$\times$10 dataset pairs. Then, we select the optimal Source Selector which performs the best (in terms of accuracy for the binary classification) on these pairs.
Finally, for each dataset in the testing group, 
we evaluate the Source Selector on the dataset pairs formed by pairing each training group dataset with the testing group dataset, in terms of accuracy. 
We denote the source selector in Smart as \texttt{Default}. We evaluate \texttt{Default} against various baselines in the following sections to reveal different insights.

\subsubsection{\textbf{Is the cross attention useful?}} 
To answer the question, we evaluate two variants of \texttt{Default}, i) \texttt{MLP} which replaces all the attention layers with a two-layer MLP and ii) \texttt{SelfAtt} which replaces the cross-attention layer with a self-attention layer.
Table~\ref{tab:sourceselector} shows the performance of the three Source Selectors on the pairs formed with each dataset in the testing groups. 
\begin{table}[t]\footnotesize
    \centering
    \begin{tabular}{ p{2.7cm}<{\centering}| p{1.41cm}<{\centering}| p{1.41cm}<{\centering}| p{1.41cm}<{\centering} }
    \hline
    \hline
    Method & \texttt{Default} & \texttt{MLP} & \texttt{SelfAtt}\\
    \hline
    \hline
    WormsTwoClass & 0.723(47/65) & 0.554(36/65) & 0.692(45/65) \\ 
    \cline{1-4}
    ItalyPowerDemand & 0.815(53/65) & 0.600(39/65) & 0.785(51/65) \\ 
    \cline{1-4}
    Lightning2 & 0.646(42/65) & 0.600(39/65) & 0.615(40/65) \\ 
    \cline{1-4}
    StarLightCurves & 0.769(50/65) & 0.708(46/65) & 0.754(49/65) \\ 
    \cline{1-4}
    Mallat & 0.800(52/65) & 0.692(45/65) & 0.754(49/65) \\ 
    \cline{1-4}
    Beef & 0.923(60/65) & 0.738(48/65) & 0.892(58/65) \\ 
    \cline{1-4}
    Ham & 0.846(55/65) & 0.754(49/65) & 0.846(55/65) \\ 
    \cline{1-4}
    Meat & 0.692(45/65) & 0.615(40/65) & 0.662(43/65) \\ 
    \cline{1-4}
    InsectWingbeatSound & 0.754(49/65) & 0.600(39/65) & 0.738(48/65) \\ 
    \cline{1-4}
    Strawberry & 0.815(53/65) & 0.769(50/65) & 0.785(51/65) \\ 
    \hline
    \end{tabular}
    \caption{Performance of Source Selectors}
    \label{tab:sourceselector}
\end{table}
As shown, \texttt{MLP} performs the worst, which suggests the advantage of applying attention mechanisms in Source Selector.
Also, \texttt{Default} consistently outperforms (or at least match) both \texttt{MLP} and \texttt{SelfAtt}, which validates the benefits of the cross-attention mechanism to capture the correlations among the heads of both datasets for source selection. 
In addition, we conduct a 17-fold cross-validation for source selection (in terms of the average accuracy), where we randomly divide all the datasets into 17 groups, pick one group as the testing group, two groups for validation, and the rest for training in each fold. Table~\ref{tab:sourceselector} show that \texttt{Default} outperforms the other two, validating our insights above.

\begin{table}[t]\footnotesize
    \centering
    \begin{tabular}{ p{1.2cm}<{\centering}| p{2cm}<{\centering}| p{2cm}<{\centering}| p{2cm}<{\centering} }
    \hline
    \hline
    Method & Default & MLP & SelfAtt\\
    \hline
    \hline
    Avg. Acc. & \textbf{0.780(272.8/350)} & 0.573(200.7/350) & 0.709(248.2/350)     \\
    \hline
    \end{tabular}
    \caption{Different Attention Designs in Source Selector}
    \label{tab:sourceselectorrepeat}
\end{table}

\begin{table}[t]\footnotesize
    \centering
    \begin{tabular}{ p{0.75cm}<{\centering}| p{0.45cm}<{\centering}| p{0.7cm}<{\centering}| p{0.7cm}<{\centering}| p{0.5cm}<{\centering}| p{0.5cm}<{\centering}| p{0.5cm}<{\centering}| p{0.5cm}<{\centering}| p{0.5cm}<{\centering}}
    \hline
    \hline
    Method &  Def. & RTS-L & RTS-S & EL-2 &EL-3 &EL-4& EL-5 & EL-6 \\
    \hline
    \hline
    Indiv. & \textbf{0.780} & 0.598 & 0.637 & 0.762 & 0.754 & 0.727 & 0.704 & 0.699\\
    \hline
    Accum. & \textbf{0.780} & N/A & N/A & 0.775 & 0.761 & 0.746 & 0.734 & 0.718 \\
    \hline
    \end{tabular}
    \caption{Different Dataset Representatives in Source Selector}
    \label{tab:sourceselectorrepeat}
    
\end{table}

\subsubsection{\textbf{Are the first-layer encoder parameters a good representation of a time series dataset?}}
By default in SMart, the first layer encoder parameters in the pre-trained TM are used as the representative of each time series dataset.
One intuitive alternative is to exploit a representative time series (RTS) in a dataset to represent the dataset, as introduced by \cite{fawaz2018transfer}. 
To exploit the RTS as the input of source selector, the source selector need to adapt accordingly. We explore the following two adaptions: i) RTS-L with an LSTM structure as the source selector. Given the $RTS_1$ for a candidate source dataset and $RTS_2$ for the target dataset, LSTM takes the following as the input: ["Source-start" token, $RTS_1$, "Target-start" token, $RTS_2$], and feed the output in the last time point to a FFN (with a softmax activation) to predict whether the candidate source dataset (that generates $RTS_1$) is beneficial for the representation learning for the target dataset (that generates $RTS_2$). ii) RTS-S with two multi-head self-attention layers, a concatenation layer and a 2-layer FFN (with a softmax activation) as the source selector. Table~\ref{tab:sourceselectorrepeat} shows the average accuracy achieved by RTS-L and RTS-S, both lower than 0.780 (by \texttt{Default}), suggesting that exploiting only a representative time series as the representation of a time series does not work as well as the first-layer encoder parameters.

Following the idea of using encoder parameters as the dataset representation, other layers' parameters may also work well. Thus, we evaluate the following variants of \texttt{Default}, by using parameters of other encoding layers (ELs), denoted by EL-$k$ where $k$ refers to the $k$-th layer. 
To be comprehensive, we test two versions: a) an individual version EL-$k$ that uses exactly the $k$-th layer's parameters to represent the dataset and b) accumulated version EL-$k$ that uses all the first $k$ layers' parameters as the dataset representation. The corresponding results are shown in Table~\ref{tab:sourceselectorrepeat}. We observe that as $k$ increases, the individual version EL-$k$ performs worse, suggesting that the later layer's parameters is worse than early layers when serving as the dataset representation. In addition, the accumulated version of EL-$k$ performs worse when $k$ increases, which suggests that the parameters in the later encoder layers do not improve over the first-layer parameters to form better dataset representation.

\subsection{SMart on Uni-variate Time Seires}
In this section, we evaluate the SMart framework on uni-variate time series datasets against the state of the arts for classification and regression.
Moreover, we conduct an ablation study to show the impacts of different components in SMart. 
Finally, we detail the impacts of challenging recovery and comprehensive recovery.

\subsubsection{Experiment Setup}
To evaluate SMart, we divide all 85 bake-off datasets into 17 groups as discussed earlier and train a Source Selector using only the training groups.
Then, for each target dataset in the testing group, we select its source datasets with the learned Source Selector to pre-train the time series model (TM; see Section 4.1). 
Note that different datasets may have time series of different lengths, leading to different lengths of the attention layer. 
To address the issue, we extend all the time series to the maximum length of all time series data, and use a padding mask (introduced in ~\cite{zerveas2021transformer}) to mask the extended part of the time series.
After then, for each downstream time series application, we connect the pre-trained encoder to an application decoder, i.e., a two-layer feed-forward net with a softmax layer for classification and a linear layer for regression,
and fine-tune the model with the training set of the target dataset (the learning rate of the encoder is 0.00001 and that of the application decoder is 0.001).
Finally, we evaluate the models on the testing set of the target dataset.

For the optimal parameter setting of TM, we have the number of encoding layers ($K$) be 6, the model dimensionality in the transformer encoding layers be 64, and the number of heads $H$ be 8. 
For the Source Selector, we set the dimensionality of head embeddings (i.e., $\{{e^\prime}^A_h\}$ and $\{{e^\prime}^B_h\}$) as 64, the model dimensionality as 128, the number of heads of the multi-head self-attention layer as 8, and the output dimensionalities of the two-layer feed-forward network as 64 and 2 (for binary classification), respectively.

\noindent
\textbf{Baselines}. We evaluate SMart against a number of state-of-the-art models for time series classification and regression (including \textbf{ROCKET}~\cite{dempster2020rocket}, \textbf{Rel-CNN}~\cite{9809831}, \textbf{TimeNet}~\cite{malhotra2017timenet}) and time series representation learning and transfer (including \noindent \textbf{TS-TCC}~\cite{eldele2021time}, \textbf{TS2Vec}~\cite{yue2022ts2vec}, \noindent \textbf{DTW-FCN}~\cite{fawaz2018transfer} and \textbf{TST}~\cite{zerveas2021transformer}).

We evaluate the proposed SMart framework which uses the cross-attention based Source Selector and recovers recurrence plots in the RPW mode (which performs the best among all three modes). We evaluate the SMarts which recover recurrence plots in RATE and RS modes in ablation study later.

\subsubsection{\textbf{Evaluation against the state of the arts.}}

\begin{figure}[!t]
\centering
\includegraphics[width=3.3in]{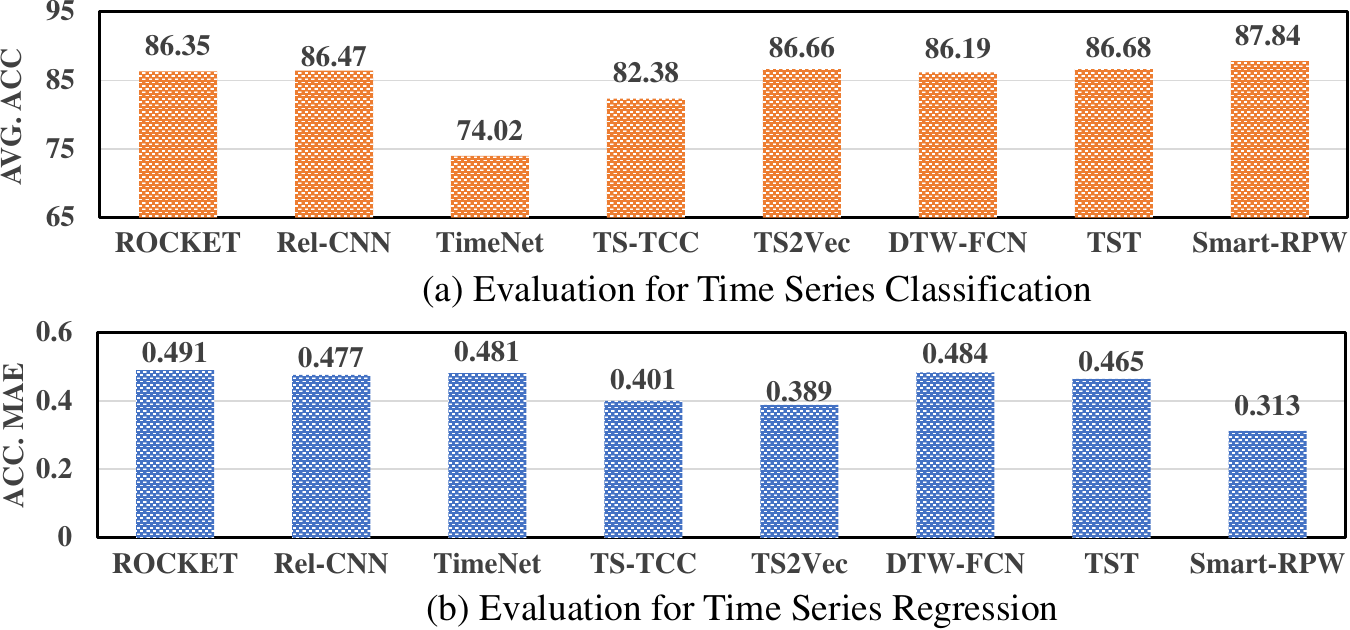}    
\caption{Evaluation on Uni-variate Time series}
\label{fig:claandreg}
\end{figure}

Figure~\ref{fig:claandreg}(a) shows the performance of all the frameworks in terms of accuracy for time series classification.
First, we observe that TimeNet performs the worst among all the models which may suggest that RNN is not a good choice of the backbone model for TSRL.
By comparing the supervised end-to-end models (i.e., ROCKET and Rel-CNN) and representation learning based frameworks (TS-TCC, TS2Vec, TST and DTW-FCN), we observe that TS2Vec and TST perform better, which suggests the effectiveness of learning time series representations for time series classification. 
Besides, we observe that DTW-FCN performs worse than ROCKET and Rel-CNN, which may be due to the base model used in DTW-FCN (i.e., a fully convolution net of fixed kernel sizes) is not competitive against ROCKET (with random kernels) and Rel-CNN (with relationship features captured).
By comparing SMart with the two most competitive baseline models (i.e., TST and TS2Vec), we observe that SMart-RPW (87.84\%) outperforms TST (86.68\%) and TS-2Vec (86.66\%), which validates the ideas of RP-recovery task and multi-source representation transfer for time series classification.
In addition, Figure~\ref{fig:claandreg}(b) shows the performance of all the frameworks in terms of mean absolute error (MAE) for regression. We have similar observations as those for classification, validating our insights again.
Finally, although we do not target on improving the training efficiency in SMart, the training time spent for each compared methods are listed in Table~\ref{tab:trainingeff}. We observe that the training time of SMart, ROCKET, TST and Rel-CNN are in the same order of magnitude.
\begin{table}[t]\footnotesize
    \centering
    \begin{tabular}{ p{1cm}<{\centering}| p{1.4cm}<{\centering}| p{1.4cm}<{\centering}| p{1.4cm}<{\centering}| p{1.4cm}<{\centering}}
    \hline
    \hline
    Method &  ROCKET & Rel-CNN & TimeNet & TS-TCC \\
    \hline
    \hline
    Time(h) & 24.5 & 21.7 & 15.1 & 2.9\\
    \hline
    Method & TS2Vec & TST & DTW-FCN & SMart\\
    \hline
    Time(h) & 1.4 & 22.5 & 28.4 & 26.6  \\
    \hline
    \end{tabular}
    \caption{Total Training Time for Compared Models}
    
    \label{tab:trainingeff}
\end{table}

\subsubsection{\textbf{Ablation Study}}
Here we perform an ablaton study to analyze the impacts of different components in SMart.
We characterize the compared variants of SMart in two aspects: A) which pre-training tasks are used? In this aspect, a framework may be i) \texttt{SELF} with self-recovery task only; ii) \texttt{RATE} that additionally recovers the recurrence plots (RPs) in the Recover-At-The-End mode; iii) \texttt{RS} that additionally recovers RPs in the Recover-in-Sequence mode; iv) \texttt{RPW}, which additionally recovers RPs in the RPW mode. 
B) how to select the source datasets? In this aspect, a framework may be i) \texttt{NoS} that only uses the target dataset to pre-train the encoder; ii) \texttt{All} that exploits all the 65 datasets in the training groups to pre-train the encoder; iii) \texttt{Type} that uses the datasets with the same data type (a tag provided by the UCR/UEA archive) as the target dataset to pre-train the encoder; iv) \texttt{DTW} that exploits the DTW distance between two datasets (defined in \cite{fawaz2018transfer}) to select the dataset most similar with the target dataset to pre-train the encoder; v) \texttt{RTS-S} that exploits a self-attention based Source Selector with the representative time series (RTS) as the input for source selection; and vi) \texttt{CrossAtt} that exploits a cross-attention based Source Selector for source selection.
Finally, we build variants of SMart by combining the design of pre-training task and the design of source selection. 
With CrossAtt as the default Source Selector, we use SMart-RATE, SMart-RS and Smart-RPW to refer to the combination of RATE, RS and RPW with CrossAtt, respectively.

\begin{table}[t]\footnotesize
    \centering
    \begin{tabular}{p{1.15cm}<{\centering}|  p{0.55cm}<{\centering}| p{1.1cm}<{\centering} | p{1.1cm}<{\centering} | p{1.1cm}<{\centering} | p{1.1cm}<{\centering} 
    }
    \hline
    \hline
    \multicolumn{1}{c}{Variants}&\multicolumn{1}{|c|}{Metric}&\multicolumn{1}{c|}{\texttt{SELF(w/o RP)}}&\multicolumn{1}{c|}{\texttt{RATE}}&\multicolumn{1}{c|}{\texttt{RS}}&\multicolumn{1}{c}{\texttt{RPW}}\\
    \hline
    \hline

    \multirow{2}{*}{\texttt{NoS}}    
     &  ACC. & 86.68  & 86.76  & 86.87 & 87.15 \\
     \cline{2-6}
     & MAE & 0.465 & 0.460 & 0.462 & 0.461 \\ 
    \hline
    \multirow{2}{*}{\texttt{All}}
    & ACC. & 83.20 & 84.27 & 84.55 & 85.63 \\ 
    \cline{2-6}
    & MAE & 0.439 & 0.410 & 0.413 & 0.405 \\ 
    \hline
    \multirow{2}{*}{\texttt{Type}}
    & ACC. & 86.53 & 86.85 & 86.89 & 86.98 \\  
    \cline{2-6}
    & MAE & 0.420 & 0.401 & 0.408 & 0.399 \\  
    \hline
    \multirow{2}{*}{\texttt{DTW}}
    & ACC. & 86.71 & 86.99 & 87.02 & 87.20 \\  
    \cline{2-6}
    & MAE & 0.416 & 0.388 & 0.395 & 0.385 \\ 
    \hline
    \multirow{2}{*}{\texttt{RTS-S}}
    & ACC. & 87.14 & 87.40  & 87.50 & 87.55 \\  
    \cline{2-6}
    & MAE &  0.374	& 0.332  &  0.336  &  0.328	 \\ 
    \hline
    \multirow{2}{*}{\texttt{CrossAtt}}
    & ACC. & 87.35 & 87.65 & 87.66 & \textbf{87.84} \\ 
    \cline{2-6}
    & MAE & 0.357 & 0.315 & 0.326 & \textbf{0.313} \\ 
    \hline
    \end{tabular}
    \caption{Ablation Study of SMart}
    \label{tab:ablation}
\end{table}

Table~\ref{tab:ablation} show the evaluation results for all the variants on classification and regression, respectively. 
As shown, combining \texttt{RPW} and \texttt{CrossAtt} (i.e., SMart-RPW) outperforms all the variants consistently, validating the effectiveness of the RPW mode for RP-recovery and the cross-attention in Source Selector.
Especially, by comparing ``CrossAtt+RPW'' and ``NoS+RPW'', we can tell the importance of exploiting the cross-attention based Source Selector for representation transfer in SMart.
Between ``CrossAtt+RPW'' and ``CrossAtt+SELF'', we observe the positive impact brought by the RP-recovery tasks on SMart.
In addition, we observe that \texttt{All} performs the worst among all the source selection methods, suggesting that it is not a good idea to simply exploit all the time series data to pre-train the encoder. 
Besides, we observe that \texttt{Type} performs better than \texttt{All}, indicating that the same type with the target dataset is a good heuristic for source selection. 
We can also compare ``DTW+SELF'' with ``DTW-FCN'' to validate the superiority of the transformer over FCN as a backbone. 
Finally, we observe that \texttt{CrossAtt} outperforms \texttt{RTS-S} in both time series classification and regression tasks, re-validating that the encoder parameters perform better than the representative time series when serving as the input to source selection for TSRL.

\subsubsection{\textbf{Impacts of Challenging Recovery (CHR) and Comprehensive Recovery (COR) in RP-Recovery.}}
From Table~\ref{tab:ablation}, we observe that RS (with only CHR) outperforms RATE for classification, while RATE (where the last representation has both CHR and COR) outperforms RS for regression no matter which source selection method is adopted. 
To explain this phenomenon, we propose a hypothesis: COR help representations to embed detailed time series status recurrence in RPs with higher granularity to benefit \texttt{RATE} for regression, while CHR in RS guide the representations produced by higher encoding layers to gradually learn the larger-scale and more precise time series dynamics better for classification.

\begin{figure}[t]
\centering
\includegraphics[width=3.35in]{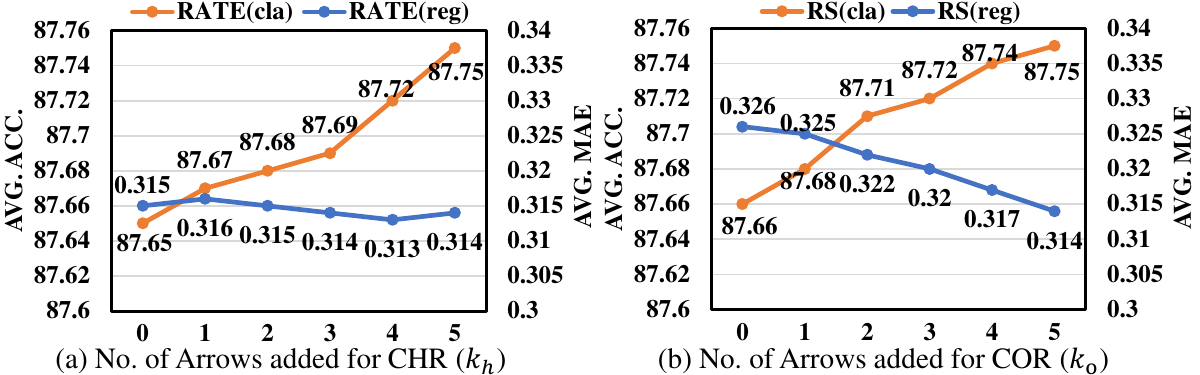}
\centering
\caption{\label{fig:arrowimpact} Evaluation for Impacts of CHR and COR}
\end{figure}

To validate the hypothesis, we control the addition of arrows (see Figure~\ref{fig:modes}) to the RP-recovery module to show the impacts of CHR and COR.
Specifically, starting from \texttt{RATE}, we add red arrows for CHR, i.e., the arrow from $\{z^{K-1}_t\}$ to $RP'_{K-1}$, the arrow from $\{z^{K-2}_t\}$ to $RP'_{K-2}$, ..., till the red arrow from $\{z^{1}_t\}$ to $RP'_{1}$, one by one. We denote the number of added CHRs as $k_h$ (\texttt{RATE} is the case with $k_h=0$) and
we set $K = 6$. Figure~\ref{fig:arrowimpact}(a) shows the performance for classification (orange line) and regression (blue line) with different $k_h$. We observe that as $k_h$ increases, the accuracy increases from 87.65\% to 87.75\% (this is not small compared to the baselines).
Meanwhile, the MAE for regression drops from 0.315 to 0.314, which is not obvious. 
The results suggest that CHRs, which guide the final representations to capture the lower-granularity (but more precise) time series dynamics better, mainly benefit time series classification.
Similarly, starting from RS, we add $(K-1)$ blue arrows for COR one by one, i.e., the arrow from $\{z^K_t\}$ to $RP'_{K-1}$, the arrow from $\{z^K_t\}$ to $RP'_{K-2}$, ..., till the arrow from $\{z^K_t\}$ to $RP'_{1}$. We use $k_o$ to denote the number of arrows added.
As shown in Figure~\ref{fig:arrowimpact}(b), a larger $k_o$ increases the accuracy for classification and decreases the MAE for regression, suggesting that COR helps representations to capture more detailed higher-granularity time series dynamics in lower RPs to benefit both classification and regression.

\subsection{Smart for Multi-variate Time Series}
In this section, we collect 9 UEA multi-variate time series datasets to evaluate SMart.
Considering that 9 datasets may not be sufficient to learn a good Source Selector, we borrow the 85 uni-variate datasets to train the Source Selector.
An intuitive question is ``whether the Source Selector learned with only uni-variate time series suitable for representation transfer among multi-variate datasets?'' If not, how to improve it?
To answer the question, we randomly divide the 9 multi-variate datasets into 3 groups, each of which includes 3 datasets, for potential training, validation and testing, respectively. Specifically, we have three plans to train the Source Selector: 1) we learn the Source Selector with uni-variate datasets only; 
2) we also exploit the validation group of multi-variate datasets for validation to select the Source Selector;
and 3) in addition to plan 2, we exploit the training group multi-variate time series datasets to form dataset pairs as well to train the Source Selector.
Finally, we evaluate the SMart with the Source Selector trained in the three plans, respectively, on the testing group of multi-variate datasets for classification and regression, respectively.

\begin{figure}[!t]
\centering
\includegraphics[width=3.45in]{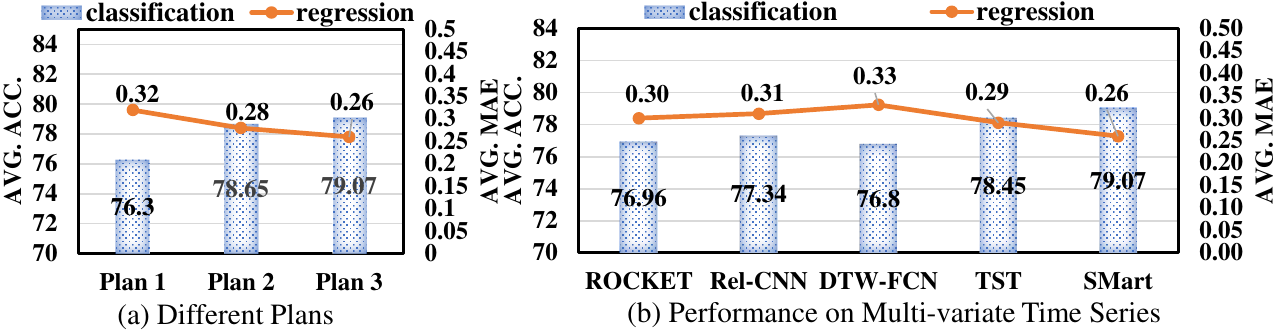}
\caption{\label{fig:multi-variate} Evaluation for Multi-variate Time Series}
\end{figure}

As shown in Figure~\ref{fig:multi-variate}(a), plan 3 performs the best (79.07\% in terms of accuracy for classification and 0.26 in terms of MAE for regression) among all the three plans, which supports the idea of including the multi-variate time series in the training group. 
In addition, we observe that SMart in plan 2 outperforms that in plan 1 for both two tasks, suggesting that including multi-variate time series for validation also benefits the source selection. 
Finally, Figure~\ref{fig:multi-variate}(b) shows the comparison among SMart and the baseline frameworks on the multi-variate time series datasets for classification and regression. The experimental results show that SMart outperforms the baselines consistently in both downstream tasks, validating the effectiveness of the representations learned by SMart.

\section{Conclusion}\label{Conclusion}
In this paper, we study the problem of time series representation learning and transfer. We propose the SMart framework which incorporates 1) an RP-recovery task as part of encoder pre-training to improve the quality of time series representation learning; and 2)
a Source Selector to select proper source datasets for pre-training the time series encoder for representation transfer. We empirically evaluate the SMart framework on both uni-variate and multi-variate datasets to validate its effectiveness. As for the next step, we plan to explore more neural network structures for Source Selector to eliminate the time complexity without compromising the performance. Also, we plan to train Source Selector with more multi-variate time series datasets.

\bibliographystyle{ACM-Reference-Format}
\bibliography{citation}

\end{document}